\documentclass{article}

\PassOptionsToPackage{numbers, compress}{natbib}

\usepackage[preprint]{neurips_2026}

\usepackage[utf8]{inputenc}
\usepackage[T1]{fontenc}
\usepackage{hyperref}
\usepackage{url}
\usepackage{booktabs}
\usepackage{amsfonts}
\usepackage{amsmath}
\usepackage{nicefrac}
\usepackage{microtype}
\usepackage{xcolor}
\usepackage{xspace}
\usepackage{bbm}

\usepackage{algorithmic}
\usepackage[ruled,vlined,linesnumbered]{algorithm2e}
\usepackage{mathtools}
\usepackage{amsthm}
\usepackage{float}
\usepackage{enumitem}

\usepackage{tabularx}
\usepackage{colortbl}
\usepackage{array}
\usepackage{listings}

\definecolor{placeholder}{RGB}{180,0,0}

\newcommand{\sip}[1]{\textsc{#1}}

\title{Counterfactual Trace Auditing of LLM Agent Skills
}

\author{%
  \textbf{Xiaolin Zhou$^{1}$ \quad Jinbo Liu$^{1}$ \quad Li Li$^{2}$ \quad
  Ryan A. Rossi$^{3}$ \quad Xiyang Hu$^{1}$} \\
  \\
  $^{1}$Arizona State University \quad
  $^{2}$University of Southern California \quad
  $^{3}$Adobe Research \\
  \texttt{\{xzhou226,\,xiyanghu\}@asu.edu}
}

\begin{document}

\maketitle

\begin{abstract}
    Large Language Model agents are increasingly augmented with \emph{agent skills}. Current evaluation methods for skills remain limited. Most deployed benchmarks report only pass rate before and after a skill is attached, treating the skill as a black box change to agent behavior. We introduce \textbf{Counterfactual Trace Auditing (CTA)}, a framework for measuring how a skill changes agent behavior. CTA pairs each \emph{with skill} agent trace with a \emph{without skill} counterpart on the same task, segments both traces into goal directed phases, aligns the phases, and emits structured \emph{Skill Influence Pattern} (SIP) annotations. These annotations describe the behavioral effect of a skill rather than only its task outcome. We instantiate CTA on SWE-Skills-Bench with Claude across 49 software engineering tasks. The resulting audit reveals a clear evaluation gap. Pass rate changes by only $+0.3$ percentage points on average, suggesting little aggregate effect. Yet CTA identifies 522 SIP instances across the same paired traces, showing that the skills substantially reshape agent behavior even when pass rate is nearly unchanged. The audit also separates several recurring effects that pass rate cannot detect, including literal template copying, off task artifact creation, excess planning, and task recovery. Three findings emerge. First, high baseline tasks contain most of the observed skill effects, although their pass rate is already saturated and therefore cannot reflect those effects. Second, tasks with moderate baseline performance show the most recoverable gain, but often at substantially higher token cost. Third, the dominant SIP type can be identified by baseline bucket: surface anchoring is most common on ceiling tasks and edge-case prompting is most common on mid-range and floor tasks. These regularities turn informal failure mode observations into reproducible behavioral measurements. Code and data are available (anonymized for review) at \url{https://github.com/WillChow66/CTA.git}.
\end{abstract}

\section{Introduction}
\label{sec:intro}

Agents built on large language models (LLMs) now ship with attachable \emph{skills}
: short Markdown documents that encode procedural knowledge for a particular framework, library, or workflow. These skills can change how an agent searches, edits, tests, and reasons. Yet the evaluation toolkit for skills remains narrow. General coding agents are usually evaluated through aggregate task success \citep{jimenez2024swebench, swebenchverified}. Skill evaluations often reduce the comparison to one scalar: the change in unit test pass rate $\Delta P$ between the \emph{with skill} and \emph{without skill} conditions on the same task. \emph{SWE-Skills-Bench} \citep{han2026sweskillsbenchagentskillsactually}, the only public benchmark we are aware of that releases paired traces for this setting, reports results in this form and also informally lists selected failure modes (surface anchoring, hallucination, concept bleed) when pass rate alone is not explanatory. This setup is useful, but it treats a skill as a black box intervention and discards most of the behavioral evidence in the trace.

The pass-rate framing has a well-known weakness: \emph{ceiling effects}. In our experiments on SWE-Skills-Bench with Sonnet 4.5, 37 of 49 tasks have $\geq 90\%$ baseline pass rate, leaving almost no headroom for a skill to register as positive $\Delta P$. But pass-rate has a less-discussed weakness as well: \emph{behavioral changes can offset}. A skill can simultaneously help an agent disambiguate the right file (constructive) and prompt it to write an extra, off-task configuration file (destructive); when both occur, $\Delta P$ may stay at zero even though the agent's trajectory has been reshaped along multiple measurable axes. We propose \textbf{Counterfactual Trace Auditing (CTA)} to address this gap by stepping inside the trace. For each task, CTA compares the \emph{with skill} run against the \emph{without skill} run from the same model, aligns their trajectories, and records where they differ in reads, writes, searches, executions, and reasoning. Because these records are attached to specific phases and intents, CTA can measure skill effects even when pass rate is saturated or when helpful and harmful effects cancel in the final outcome.

CTA operates on a \emph{paired trace bundle} for each task: the event stream produced with the skill, the event stream produced without the skill, the skill document, and the evaluation reports for the two resulting repositories. The pipeline parses each trace into typed events ($\textsc{read}$, $\textsc{write}$, $\textsc{execute}$, $\textsc{search}$, $\textsc{think}$), segments each trace into five goal directed phases ($\textsc{Orientation}$, $\textsc{Implementation}$, $\textsc{Validation}$, $\textsc{Debugging}$, $\textsc{Finalization}$) using a deterministic finite state machine, aligns the two traces at the phase level using dynamic time warping, aligns actions within each phase at the intent level using TF-IDF cosine similarity over reasoning text, and emits a \emph{divergence record} for every aligned pair whose action windows differ. A separate recovery step records \emph{unilateral} actions, such as writes performed by the with skill agent on files never touched by the without skill agent. CTA then maps structural divergence records to \emph{Skill Influence Patterns} (SIPs).


We instantiate CTA on the full public \emph{SWE-Skills-Bench} corpus with Claude Sonnet 4.5, using 49 tasks and two traces per task. The audit reveals a clear evaluation gap. Mean $\Delta P$ is only $+0.3$ percentage points, but CTA identifies 522 SIP instances across the same paired traces. High baseline tasks contain most observed skill effects despite little pass rate headroom. Mid baseline tasks carry the main recoverable gains, but often with much higher token cost. Low baseline tasks more often show \sip{Edge Case Prompting} without successful repair. The dominant SIP type can also be identified by baseline level: \sip{Surface Anchoring} is most common on high-baseline tasks, and \sip{Edge Case Prompting} is most common on mid- and low-baseline tasks. These results show that skill effects are structured, measurable, and often invisible to pass rate alone.
We complement the aggregate audit with several mechanism case studies (\S\ref{sec:cases}). These cases show when a skill produces recoverable improvement, when it spends extra tokens without measurable gain, and when it causes premature closure. The premature closure case is important because it causes the corpus's only negative $\Delta P$ event without firing a harmful SIP, marking a concrete limit of the current taxonomy.

To our knowledge, CTA is the first released framework focused on trace-level auditing of paired with-skill / without-skill software-engineering agent trajectories. We contribute:
\begin{itemize}[leftmargin=*, itemsep=0pt]
    \item the \textbf{CTA framework} (\S\ref{sec:framework}): a pipeline that pairs \emph{with skill} and \emph{without skill} traces for the same task and model, segments them into goal directed phases, aligns them at the phase and intent levels, and emits structured divergence records, including unilateral writes that alignment would miss;
    \item a \textbf{five class SIP taxonomy} (\S\ref{sec:sips}) with deterministic rule based detectors for constructive, neutral, and destructive skill effects;
    \item a \textbf{49 task observational study} (\S\ref{sec:results}) on the public \emph{SWE-Skills-Bench} with Claude Sonnet 4.5 that quantifies the gap between $\Delta P$ and structural behavior change, and shows that the dominant SIP type can be identified by baseline bucket;
    \item \textbf{mechanism case studies} (\S\ref{sec:cases}) on tasks with large $|\Delta P|$, including a premature closure case that fires no harmful SIP and motivates a future taxonomy extension.
\end{itemize}

\section{Related Work}
\label{sec:related}

\textbf{Agent skills.} Agent skills are document-form interventions that add task specific procedures, templates, examples, and code snippets to an agent context at run time. Current agent platforms make it possible for users and third parties to define such skills as Markdown artifacts and invoke them conditionally for a task \citep{anthropic2025skills}. In research, this design is closely related to retrieval-augmented and tool-augmented agents \citep{li2026defensespromptattackslearn}, where the model selects external information or actions as part of its trajectory \citep{schick2023toolformer, yao2023react, li2026autonomytaxdefensetraining}. Existing skill evaluations usually ask whether attaching the document changes final task success. For example, \emph{SWE-Skills-Bench} \citep{han2026sweskillsbenchagentskillsactually} reports per skill $\Delta P$ on unit tests and gives qualitative notes on several failure modes. Additionally, \emph{SkillsBench}\citep{li2026skillsbench} claims that self-generated skills provide no average benefit across diverse tasks, while using 2-3 focused skills outperform comprehensive documentation. \emph{SkillTester} \citep{wang2026skilltester}, closely related to our framework, proposes a paired baseline / with-skill harness, however, it normalizes the contrast into a single score rather than localizing changes within the trace. 
In contrast to previous literature, CTA studies a different object: the paired behavioral trace. It asks which actions, phases, and artifacts changed after using the skill.

\textbf{Agent benchmarks for code.} Code agent benchmarks such as \emph{SWE-Bench} \citep{jimenez2024swebench}, \emph{SWE-Bench Verified} \citep{swebenchverified} evaluate an agent by whether its generated patch passes tests. This objective is appropriate for measuring end to end repair performance, but it is not designed to isolate the effect of an attached skill. In particular, a pass rate delta does not show whether the skill changed file search, patch selection, validation, debugging, or finalization.
To address the limitation, a growing line of work moves from pass-rate to process: \citet{chen2025beyond} analyze SWE-bench trajectories to identify execution-error patterns that pass-rate alone cannot expose, and \citet{mehtiyev2026beyond} show that behavioral patterns in the trajectory, beyond aggregate outcome metrics, drive coding agent success and failure. These works confirm that process-level signals matter, but they study agent capability in general rather than the effect of using skills.
\emph{SWE-Skills-Bench} \citep{han2026sweskillsbenchagentskillsactually} is the closest prior benchmark for our setting because it releases paired \emph{with skill} and \emph{without skill} traces. We use that corpus, but replace selected qualitative failure notes with a general trace auditing pipeline: phase segmentation, intent level alignment, divergence records, and deterministic SIP detectors.

\textbf{Agent trajectory analysis.} Another line of work studies the trajectory of an LLM system rather than only its final output. Reflexion \citep{shinn2023reflexion} and Self Refine \citep{madaan2023selfrefine} use prior steps as material for revision. AutoGen \citep{wu2024autogen} records multi agent message logs to support debugging and coordination analysis. More directly relevant to our setting, TRACE~\citep{kim2025beyond} evaluates tool-augmented agent trajectories along multiple dimensions beyond final-answer matching, and SWE-PRM~\citep{gandhi2025agents} identifies recurring trajectory-level errors such as redundant exploration, looping, and failure to terminate \citep{li-etal-2025-treble}. However, they usually analyze one trajectory at a time. CTA instead compares two trajectories for the same task and the same base agent, one with the skill and one without it. This pairing lets us define a divergence as a contrast between behaviors, rather than as an absolute property of a single run. 

\textbf{In context anchoring and instruction following.} LLM behavior depends strongly on the format, order, and placement of contextual information. \citep{zhao2021calibrate,limm,Li_Ji_Wu_Li_Qin_Wei_Zimmermann_2024} show majority label and recency biases from in context examples. \citep{lu2022fantastically} find that example ordering alone can change accuracy by large margins. \citep{wei2022chain} study chain of thought prompting as a mechanism for eliciting intermediate reasoning. \citep{liu2024lost} show that information placed in the middle of long contexts can receive less attention than information near the beginning or end. Skill injection is a structured instance of this broader setting. It does not only add facts to the prompt; it adds procedures, templates, and examples that can steer action selection \cite{Li_2025_CVPR, li2025secureondevicevideoood}. CTA connects these prompting time effects to trace level events. \sip{Surface Anchoring} marks cases where the agent copies or follows literal skill text in a way not supported by the task, building on documented \emph{copy bias}~\citep{ali2026mitigating} and token co-occurrence reinforcement effects in in-context learning~\citep{yan2024understanding}. \sip{Concept Bleed} marks cases where concepts from the skill appear in off task edits or artifacts. The key distinction is observability: CTA records where each effect occurs, which phase contains it, which action windows differ, and which skill content is matched.

\section{The CTA Framework}
\label{sec:framework}

\begin{figure}[t]
    \centering
    \includegraphics[width=\linewidth]{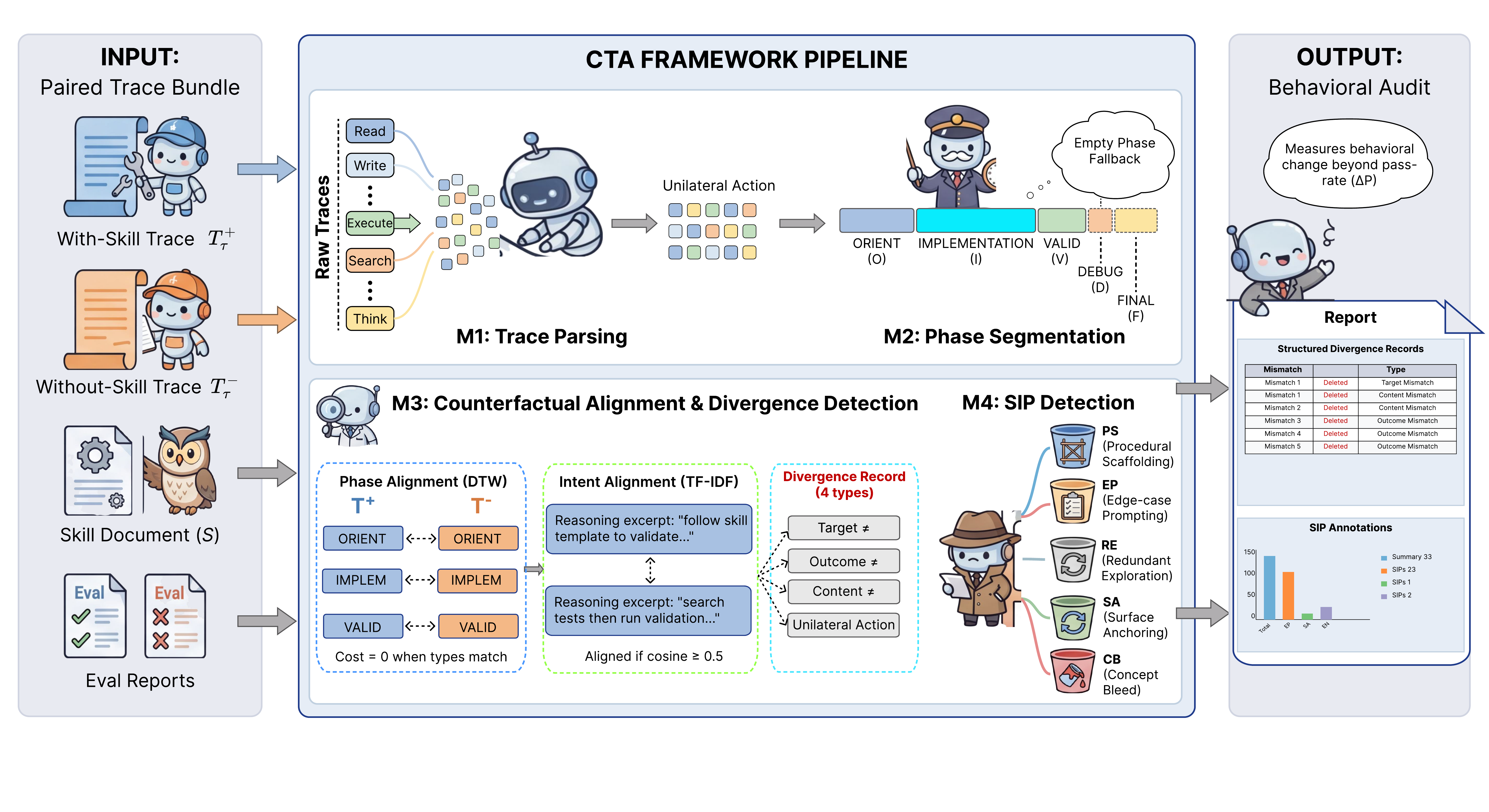}
    \vspace{-13mm}
    \caption{\textbf{Counterfactual Trace Auditing (CTA)}. For each task, CTA compares a paired set of agent trajectories generated with and without an attached skill. The pipeline parses raw logs into typed events, segments each trace into goal-directed phases using a deterministic finite state machine, aligns the two traces at the phase and intent levels, and extracts divergence records that localize behavioral differences. Each divergence is then mapped to one or more Skill Influence Patterns (SIPs), yielding a structured audit of how the skill changes agent behavior beyond aggregate pass rate.
    }
    \label{fig:cta_overview}
\end{figure}

For each task $\tau$, Counterfactual Trace Auditing (CTA) takes as input a paired trace bundle
$\mathcal{B}_\tau = \big(q_\tau,\; T^{+}_\tau,\; T^{-}_\tau,\; S_\tau,\; r^{+}_\tau,\; r^{-}_\tau\big), $
where $q_\tau$ is the task specification, $T^{+}_\tau$ is the trace produced with the skill attached, $T^{-}_\tau$ is the trace produced without the skill, $S_\tau$ is the skill document, and $r^{+}_\tau,r^{-}_\tau \in [0,1]$ are the unit test pass rates of the final repository states. The two traces are generated by the same base agent on the same task, with skill availability as the experimental condition. Figure~\ref{fig:cta_overview} gives an overview of the pipeline.

Each trace is an ordered sequence of typed events
\[
e = (t,\; \mathrm{type},\; \mathrm{reasoning},\; \mathrm{tool\_input},\; \mathrm{tool\_output}),
\]
with
$\mathrm{type} \in \{\textsc{read},\textsc{write},\textsc{execute},\textsc{search},\textsc{think}\}.$
The trace records what the agent read, wrote, searched, executed, and reasoned about during the task.

We use the term \emph{counterfactual} in an operational sense. A counterfactual pair is a matched pair of traces for the same task and same base model, one with the skill and one without it. This design controls for task identity and agent identity, but it does not control for all sources of stochastic variation, such as decoder sampling, prompt time tokenization effects, hidden platform changes, or tool scheduling. CTA therefore provides a descriptive contrast between the skill condition and the no skill condition. It is not a causal estimand in the Pearl or Rubin sense.

CTA emits two structured outputs. First, it emits a set of \emph{divergence records}. A divergence record localizes a behavioral difference between an aligned with skill window and without skill window. Each record stores the task, phase, aligned intent windows, action windows, divergence type, affected targets, and normalized features used by later detectors. We use four divergence types:
\[
\texttt{target\_mismatch},\quad
\texttt{content\_mismatch},\quad
\texttt{outcome\_mismatch},\quad
\texttt{unilateral\_action}.
\]
Second, CTA assigns zero or more Skill Influence Pattern (SIP) labels to each divergence:
$\mathcal{L}(D_k) \subseteq \{\sip{PS},\sip{EP},\sip{RE},\sip{SA},\sip{CB}\}.$
SIP scores are deterministic rule scores in $[0,1]$, not calibrated probabilities. A divergence may receive no SIP label, one SIP label, or several SIP labels.

The pipeline has four modules. M1 parses raw agent traces into typed event streams. M2 segments each trace into task phases. M3 aligns the two traces and emits divergence records. M4 maps divergence records to SIP labels using deterministic detectors.

\textbf{Phase fallback.}
The phase segmenter can return an empty phase list for traces that begin with execution rather than file inspection. This pattern occurs in shell driven tasks and test driven workflows. When this happens, CTA assigns the entire trace to a single \textsc{Implementation} phase. The fallback does not alter the event stream. It only makes the phase assignment explicit, so that such traces remain auditable rather than being dropped.

\textbf{Unilateral actions.}
A symmetric alignment can miss actions that occur only in the with skill trace. This is especially important for new files, auxiliary scripts, and template derived artifacts. CTA therefore adds a one sided recovery pass. If the with skill trace writes to a target that is never touched in the without skill trace, CTA emits a \texttt{unilateral\_action} divergence with an empty without skill action window. This is a divergence type, not a sixth SIP class. The SIP detectors may later label the same record as, for example, \sip{SA} or \sip{CB}.

\subsection{M1: Trace parsing}

M1 parses the stream JSON format used by the SWE-Skills-Bench run harness. It converts each raw message or tool event into a typed event, extracts reasoning text when present, maps tool calls to target files when possible, normalizes Unix paths, and records tool outcomes. It also computes a per trace token total from reported usage fields. The result is a uniform event stream that abstracts over low level message formatting while preserving the action sequence needed for auditing.

\subsection{M2: Phase segmentation}
\label{sec:m2}

M2 segments each trace into five phases using a deterministic finite state machine:
\[
\textsc{Orientation},\quad
\textsc{Implementation},\quad
\textsc{Validation},\quad
\textsc{Debugging},\quad
\textsc{Finalization}.
\]
\textsc{Orientation} contains initial inspection, search, and reasoning before the first write. \textsc{Implementation} contains code or artifact edits. \textsc{Validation} contains test or build executions, including commands such as \texttt{pytest}, \texttt{npm test}, \texttt{mvn test}, and \texttt{cargo test}. \textsc{Debugging} contains edits that follow a failing validation event. \textsc{Finalization} contains terminal actions after successful validation or agent exit. A trace may contain repeated \textsc{Implementation}, \textsc{Validation}, and \textsc{Debugging} spans. The segmenter is intentionally conservative: it favors fewer, larger phases over many short phases. This reduces spurious phase boundaries and leaves finer alignment to M3. 

\subsection{M3: Counterfactual alignment}

M3 aligns the with skill and without skill traces at two levels: phases and intents. Let
$\Phi^{+}_\tau = (\phi^{+}_1,\ldots,\phi^{+}_m), \Phi^{-}_\tau = (\phi^{-}_1,\ldots,\phi^{-}_n)$ 
be the phase sequences from M2. CTA first computes a phase alignment using dynamic time warping with cost
\begin{equation}
\label{eq:phase-cost}
d(\phi_i^{+}, \phi_j^{-})
=
\mathbf{1}\{
\mathrm{type}(\phi_i^{+}) \neq
\mathrm{type}(\phi_j^{-})
\}.
\end{equation}
The cost depends only on phase type. We use this type only cost because the phase segmenter is conservative, and event counts inside a phase can vary for reasons unrelated to skill influence. 

Within each aligned phase pair, CTA extracts intent windows. An intent window begins at a \textsc{think} or reasoning event and extends to the next reasoning event or phase boundary. The intent text is the concatenated reasoning string, and the tail action window is the sequence of tool events that follows it. CTA aligns intent windows using TF IDF cosine similarity over sentence level reasoning text, with threshold $\delta=0.5$. Intent pairs below threshold are treated as unaligned.

For each aligned intent pair, CTA compares the corresponding tail action windows. A divergence record is emitted when at least one of the following changes: the target differs, the target is the same but the written content differs, the action is similar but the observed outcome differs, or the action is present in only one trace. A separate recovery pass emits \texttt{unilateral\_action} records for with skill writes to targets never touched by the without skill trace.


\subsection{M4: SIP detection}
\label{sec:sips}

We classify each divergence into one of five Skill Influence Patterns. The categories were distilled from a manual reading of $\sim$50 divergences during pilot runs against the literature anchors in \S\ref{sec:related}, and refined to ensure each class has a deterministic, trace-observable signature. Three of our five SIPs (\sip{SA}, \sip{CB}, \sip{RE}) match informal failure modes already named in SWE-Skills-Bench, with our contribution being a deterministic detection signature and a release of fired instances. Two (\sip{PS}, \sip{EP}) capture constructive influence patterns that prior work conflated with ``skill helped'' without distinguishing the underlying mechanism.

\textbf{\sip{Procedural Scaffolding} (PS, constructive).} The skill provides a step sequence the agent's parametric knowledge omits (formula, protocol handshake). Signature: with-skill phase order tracks a section of $S_\tau$; key implementation events cite a skill step; without-skill omits corresponding step.

\textbf{\sip{Edge-case Prompting} (EP, constructive).} The skill acts as a checklist for easily-missed branches. Signature: with-skill writes contain extra \texttt{if/try/assert} guards on a target; without-skill touches the same target without the guard; the skill explicitly enumerates the case.

\textbf{\sip{Redundant Exploration} (RE, neutral).} The skill repeats knowledge the agent already has, or induces extra Implementation--Validation cycles that converge on the same final code. Signature: high intent-cosine but elevated action count; final diff has small AST distance; with-skill token count $\gg$ without-skill but outcome unchanged. (RE merges the original $\textsc{Redundant Reiteration}$ and $\textsc{Parallel Exploration}$ classes from the pilot taxonomy.)

\textbf{\sip{Surface Anchoring} (SA, destructive).} The agent verbatim copies a literal token from the skill template (version pin, API parameter, import path) into project code where it is incompatible. Signature: an $n$-gram ($n\ge 3$) from $S_\tau$ appears literally in a with-skill write; the same string is absent from the without-skill trace; the literal is incompatible with the project's actual dependency or config.

\textbf{\sip{Concept Bleed} (CB, destructive).} The skill's broad coverage causes the agent to introduce content not requested by the task. Signature: number of write targets is strictly larger with-skill; new targets are not in the requirement's File-Operations list; new content has high cosine to a section of $S_\tau$ that is unrelated to the requirement.

M4 maps divergence records to SIP labels using deterministic rule based detectors. Each detector consumes a divergence record, the task specification $q_\tau$, the skill document $S_\tau$, the aligned action windows, and local reasoning text. For each SIP class $\ell$, the detector returns a score $c_\ell(D_k) \in [0,1].$
The SIP fires when
$c_\ell(D_k) \geq \theta, \theta = 0.50.$
The five detectors run independently. Therefore, aggregate SIP counts are counts over divergence label pairs $(D_k,\ell)$ above threshold, not counts over divergences. This distinction matters because one divergence can express more than one skill influence pattern. For example, a with skill write may copy a literal string from the skill and create a task irrelevant file, firing both \sip{SA} and \sip{CB}. Conversely, some divergences may receive no SIP label.




\section{Experiment Setup and Findings}
\label{sec:results}

We use the public SWE-Skills-Bench corpus, where each task ships with a curated skill document $S_\tau$ targeting that task. We evaluate 49 paired bundles selected/aggregated at the skill level. We run Claude Sonnet 4.5 twice per task (once with skill, once without) under the bench-provided agent harness. We obtain 49 paired-trace bundles, 49 \texttt{eval\_report\_*.json} pass-rate files per condition, 49 task-metadata records, and 49 per-task CTA outputs. Sonnet 4.5 inference cost dominated our budget. We made an explicit trade-off: cover all 49 tasks at $r=1$ rather than the original plan's $r=3$ on a 17-task subset. We discuss the consequences for variance estimation in \S\ref{sec:limitations}. For each task we read the unit-test (L2) pass count from the eval report and compute $\Delta P_\tau = r^{+}_\tau - r^{-}_\tau$.



\textbf{Notation.} $\Delta P$ denotes a difference of two pass rates and is reported in \emph{percentage points} (pp), not percent. ``$+18.2$ pp'' means the with-skill repository passed $18.2$ pp more of the L2 unit-test items than the without-skill repository on the same task (e.g.\ $0.73 \to 0.91$); we do \emph{not} use ``$+18.2\%$'' to mean a relative gain. Token-overhead ratios remain unitless multiples (``$2.77\times$'').

\subsection{Pass-rate is nearly silent --- behavior is not}

Across all 49 tasks the mean pass-rate change is $\Delta P = +0.34$ pp, median $0$ pp, standard deviation $4.4$ pp. Only 3 tasks have $\Delta P \geq +4$ pp (\texttt{bash-defensive-patterns} $+18.2$ pp, \texttt{gitlab-ci-patterns} $+14.3$ pp, \texttt{add-admin-api-endpoint} $+4.0$ pp); 1 task is hurt ($\Delta P = -20.0$ pp on \texttt{prompt-engineering-patterns} from a $100\%$ baseline); the remaining 45 tasks have $\Delta P = 0$ pp. A reader who saw only this aggregate would conclude that the skills are essentially inert. The same 49 traces, however, contain \emph{696 behavioral divergences} between the with- and without-skill agent (mean $14.2$ per task) and \emph{522 SIP instances} (mean $10.7$ per task) --- our central observation. (Each $\Delta P$ here is a difference between two scalar L2 unit-test pass rates from a \emph{single} with-skill and a single without-skill run; we discuss the consequences of $r=1$ in \S\ref{sec:limitations}.)

\begin{table}[t]
\centering
\caption{Pass-rate $\Delta P$ versus structural divergence and SIP count, stratified by baseline pass-rate. All numbers auto-generated by \texttt{scripts/cta\_paper\_stats.py}.}
\label{tab:stratified}
\begin{tabular}{lrrrrrrrr}
\toprule
 & $n$ & Baseline & $\Delta P$ (pp) & Tok.\,$\times$ & \#Div & \#SIP & \#Div/task & \#SIP/task \\
\midrule
Ceiling  ($\geq 0.9$) & 37 & 0.98 & -0.5 & 1.64 & 531 & 415 & 14.4 & 11.2 \\
Mid      ($0.5\!-\!0.9$) & 10 & 0.69 & +3.6 & 2.77 & 141 & 92 & 14.1 & 9.2 \\
Floor    ($<0.5$) & 2 & 0.24 & +0.0 & 2.60 & 24 & 15 & 12.0 & 7.5 \\
\midrule
All & 49 & 0.89 & +0.3 & 1.91 & 696 & 522 & 14.2 & 10.7 \\
\bottomrule
\end{tabular}

\vspace{-5mm}\end{table}

\subsection{Stratification reveals a ceiling-driven evaluation gap}

Table~\ref{tab:stratified} stratifies by baseline pass rate. Three observations.

\textbf{(1)} The 37 ceiling tasks ($r^{-} \geq 0.9$) absorb 80\% of all SIP instances ($415/522$) but contribute essentially zero net pass-rate change ($-0.5$ pp on average). This is the evaluation gap: the bulk of skill-induced behavioral change happens precisely where $\Delta P$ cannot register it.

\textbf{(2)} The 10 mid-range tasks ($0.5 \leq r^{-} < 0.9$) carry the recoverable signal: $\Delta P = +3.6$ pp, $9.2$ SIPs/task, and a $2.77\times$ token cost relative to baseline. We argue that future skill benchmarks should be built on tasks in this regime rather than on tasks where pass-rate already saturates.

\textbf{(3)} The 2 floor tasks ($r^{-} < 0.5$) attract the most \sip{Edge-case-Prompting} per task ($4.5$) but no net pass-rate gain; \S\ref{sec:cases} explains why.


\subsection{Per-task SIP composition shifts with baseline}

\begin{table}[H]
\centering
\caption{Mean SIPs per task by category, stratified by baseline pass rate. Constructive: PS, EP. Neutral: RE. Destructive: SA, CB.}
\label{tab:sip-distribution}
\begin{tabular}{lrrrrr}
\toprule
 & PS & EP & RE & SA & CB \\
 & (constr.) & (constr.) & (neutral) & (destr.) & (destr.) \\
\midrule
Ceiling ($n=37$) & 0.30 & 3.97 & 0.59 & 4.38 & 1.97 \\
Mid ($n=10$) & 0.00 & 3.00 & 2.00 & 1.60 & 2.60 \\
Floor ($n=2$) & 0.00 & 4.50 & 0.00 & 3.00 & 0.00 \\
\midrule
All ($n=49$) & 0.22 & 3.80 & 0.86 & 3.76 & 2.02 \\
\bottomrule
\end{tabular}

\vspace{-5mm}\end{table}

Table~\ref{tab:sip-distribution} is the second main empirical finding: the SIP signature is not bucket-invariant. \sip{SA} dominates on ceiling tasks ($4.38$/task), where the agent already does the right thing and a literal copy from the skill is the salient deviation; \sip{EP} dominates on mid-range tasks ($3.00$/task), where the skill most often prompts additional edge-case handling; and \sip{EP} also dominates on the two floor tasks ($4.50$/task), consistent with floor tasks being underspecified relative to the skill’s checklist coverage.

\subsection{Most divergence is upfront}

$84\%$ of divergences fall in the \textsc{Orientation} ($44\%$) and \textsc{Implementation} ($40\%$) phases; \textsc{Validation}, \textsc{Debugging}, and \textsc{Finalization} together carry $16\%$. The conclusion is robust to the M2 empty-phase fallback (\S\ref{sec:framework}): when we exclude the 16 tasks that hit the fallback (and whose entire trace is collapsed to one \textsc{Implementation} phase), the \textsc{Orientation}+\textsc{Implementation} share is still $79\%$ ($57\%+22\%$), so the skill-induced signal really is concentrated in the upfront half of the trace and not an artifact of fallback-induced phase collapse. This is consistent with the hypothesis that skills change \emph{what the agent attends to} more than \emph{how it recovers from errors}: by the time the agent enters debugging, both branches typically share the same failing-test signal and converge.

\subsection{Token cost is real, heavy-tailed, and not free}

\begin{table}[H]
\centering
\caption{Six tasks with the largest $|\Delta P|$ in our 49-task sample.}
\label{tab:top-delta}
\begin{tabular}{lrrrr}
\toprule
Task & Baseline & With-skill & $\Delta P$ (pp) & Tok.\,$\times$ \\
\midrule
\texttt{prompt-engineering-patterns} & 1.00 & 0.80 & -20.0 & 1.09 \\
\texttt{bash-defensive-patterns} & 0.73 & 0.91 & +18.2 & 0.90 \\
\texttt{gitlab-ci-patterns} & 0.64 & 0.79 & +14.3 & 22.24 \\
\texttt{add-admin-api-endpoint} & 0.84 & 0.88 & +4.0 & 0.02 \\
\texttt{add-malli-schemas} & 0.90 & 0.90 & +0.0 & 0.11 \\
\texttt{add-uint-support} & 0.50 & 0.50 & +0.0 & 0.01 \\
\bottomrule
\end{tabular}

\vspace{-5mm}\end{table}

The mean token-overhead ratio is $1.91\times$ overall, with median $1.09\times$ but a long right tail (Table~\ref{tab:top-delta} reports the six tasks with the largest $|\Delta P|$): 12 ceiling tasks have token overhead $\geq 1.5\times$ at $\Delta P \leq 0$ pp, peaking at $6.80\times$ on \texttt{creating-financial-models} (a ceiling task whose pass rate did not change at all). On the mid-range tasks where skills do help, the mean is $2.77\times$, dominated by a single task (\texttt{gitlab-ci-patterns}, $22.24\times$ for a $+14.3$ pp gain). Skills are therefore not a no-op for production systems even when $\Delta P = 0$ pp, and the cost distribution is skewed enough that mean overhead alone obscures the worst cases.

\section{Mechanism Case Studies}
\label{sec:cases}
We describe five mechanisms by which skills reshape agent behavior. Consistent with our $r=1$ design, each case is grounded in the divergence/SIP records of one paired bundle (with-skill trace, without-skill trace, eval reports), with the exact release file cited inline. We do not claim that these mechanisms are statistically estimated. Rather, each is concretely measurable on the cited bundle and appears in similar form elsewhere in the corpus. The full skill excerpt and side-by-side trace diff are reproduced in Appendix~\ref{app:cases}.

\textbf{Case 1: Procedural premature-closure (negative; out-of-taxonomy).}
\texttt{prompt-engineering-patterns} is the only $\Delta P<0$ case in our 49-task corpus: $r^{-}=1.00$, $r^{+}=0.80$, $\Delta P=-20.0$ pp. The auditor records 9 divergences, all in \textsc{Implementation}, but none fires any of the 5 SIP detectors at the $0.50$ threshold. Token cost is not the issue: the with-skill trace uses only $1.09\times$ baseline tokens. The skill ends with a ``commit and document'' step, and the with-skill agent stops there, whereas the baseline continues into validation that the unit-test target depends on. Mechanism: the skill supplies an explicit completion boundary that the agent treats as terminal, suppressing its default validation loop. We therefore treat this as an out-of-taxonomy failure mode, tentatively \sip{Premature Closure}, but do not add a sixth class from a single task.

\textbf{Case 2: Search-space pruning at high token cost (positive).}
\texttt{gitlab-ci-patterns} improves from $r^{-}=0.64$ to $r^{+}=0.79$, giving $\Delta P=+14.3$ pp. The auditor records 16 divergences: 12 in \textsc{Orientation} and 4 in \textsc{Implementation}. The 13 SIP fires are dominated by \sip{RE} (6), followed by \sip{SA} (3), \sip{EP} (2), and \sip{CB} (2). Token overhead is $22.24\times$, the largest in the corpus. The trace shows the with-skill agent reading and consulting large parts of the skill while considering \texttt{.gitlab-ci.yml} layouts the baseline never explores. Mechanism: the skill reduces implementation-phase search but imposes a very large orientation-phase reading cost. The net effect is positive, although destructive \sip{SA} and \sip{CB} components remain visible in the SIP profile.

\textbf{Case 3: Surface-anchoring despite positive $\Delta P$ (mixed).}
\texttt{bash-defensive-patterns} improves from $r^{-}=0.73$ to $r^{+}=0.91$, giving $\Delta P=+18.2$ pp. The bundle has 11 divergences, all in \textsc{Implementation}, including 2 \sip{Unilateral\_Action} cases. It has 16 SIP fires, dominated by \sip{SA} (10), then \sip{CB} (3), \sip{EP} (2), and \sip{RE} (1). Token overhead is $0.90\times$, so the with-skill trace is shorter than baseline. Mechanism: the skill provides a defensive-shell template that the agent applies almost verbatim. Some literal copies are correct edge-case handling, which explains the \sip{EP} fires and the pass-rate gain, but the dominant behavior is retrieval rather than synthesis. This shows why pass rate alone can hide partly destructive dynamics.

\textbf{Case 4: Unilateral artifacts as a corpus-wide phenomenon (mixed).}
The \sip{Unilateral\_Action} pass in M3 catches with-skill writes to targets the baseline never touches. It accounts for 112 of 696 corpus-wide divergences ($16\%$). In \texttt{bash-defensive-patterns}, two such writes create test artifacts that help the unit-test outcome. Across the corpus, however, many unilateral writes are auxiliary configuration files, documentation, or extra test scaffolding not requested by the task, contributing to the \sip{CB} signal in the mid-range bucket (Table~\ref{tab:sip-distribution}, $2.60$/task). Without this pass, symmetric alignment would drop these divergences, so the pass improves detection coverage rather than merely changing annotation style.

\textbf{Case 5: Skill inertia at the ceiling (cost-only).}
At the corpus level, 12 ceiling tasks ($r^{-}\geq 0.9$) have token overhead $\geq 1.5\times$ with $\Delta P\leq 0$ pp. Thus, one third of ceiling tasks pay a sustained token tax without measured benefit. The largest cases are \texttt{creating-financial-models} ($6.80\times$, $\Delta P=0$), \texttt{spark-optimization} ($6.19\times$), \texttt{python-packaging} ($4.73\times$), and \texttt{distributed-tracing} ($4.27\times$). In these tasks, the baseline already passes all unit tests, so additional with-skill activity appears only as cost. Mechanism: skills prescribe process, while agents optimize outcome. On saturated tasks, the skill can make the agent continue after the outcome is already achieved. 

\section{Discussion}
\label{sec:discussion}

\subsection{When does $\Delta P$ undercount skill influence?}

Our results identify three regimes.
\textbf{(a)} On ceiling tasks ($r^{-}\geq 0.9$), $\Delta P$ is structurally bounded: skills can only appear as harms or token cost (Cases 1, 5). Thus, skill papers that report only $\Delta P$ in this regime report little. In \S\ref{sec:results}, the discriminative-variance comparison shows this directly: across the 36 ceiling tasks with $\Delta P=0$ pp, \#SIP std is $8.9$ and tok-ratio std is $1.60$, while $\Delta P$ std is $0$. Pass rate has no discriminative power on this sub-population, but CTA still rank-orders tasks through trace-level signals.
\textbf{(b)} On mid-range tasks, $\Delta P$ is informative but hides composition (Cases 2--4). For example, \texttt{bash-defensive-patterns} has a $+18.2$ pp gain with 10 \sip{SA} fires, while \texttt{gitlab-ci-patterns} has a $+14.3$ pp gain with $22\times$ token cost. CTA separates constructive effects from latent destructive or costly components; $\Delta P$ alone does not.
\textbf{(c)} On floor tasks ($r^{-}<0.5$, $n=2$), the skill is consumed but does not move the agent off the failing baseline trajectory. The dominant SIP is \sip{EP} ($4.5$/task), suggesting that the edge-case checklist is applied but does not address the deeper bottleneck. We omit a floor-task case study because $n=2$ is too small.

\subsection{Implications for skill design}

The dominance of \sip{SA} on ceiling tasks ($4.38$/task) and the existence of mechanism Case 5 (\emph{skill inertia}) jointly suggest a concrete design rule: \textbf{skills should describe properties of correct outputs, not procedures to produce them.} Procedural skills compete with the agent's default loop and create the very SIPs that depress $\Delta P$. Declarative ``what does correct look like'' skills are less likely to over-anchor the agent or trigger premature closure.

\section{Limitations}
\label{sec:limitations}

\textbf{Single repetition.} We run each (task, condition) once. We cannot estimate within-task variance from this design and cannot compute task-level confidence intervals on $\Delta P$. The bucket-level means in Table~\ref{tab:stratified} are descriptive cross-task summaries, not estimates of expected skill effect.

\textbf{Single model, single benchmark.} All numbers are from Claude Sonnet 4.5 on SWE-Skills-Bench. The CTA framework is model- and benchmark-agnostic by construction (the input is just a paired trace bundle), but the SIP \emph{frequencies} we report are not.

\textbf{Rule-based detector, no human gold set.} M4 is a deterministic rule ensemble, not a trained classifier validated against a human-annotated gold set. 
Building a multi-judge LLM annotation set in the style of \citep{gilardi2023chatgpt} is the natural next step but is excluded from this paper to avoid the obvious circularity (LLMs both produce and judge the traces).

\textbf{Phase segmenter not human-validated.} M2 is a hand-tuned FSM. A small number of traces fall into the empty-phase fallback (\S\ref{sec:framework}); these contributed several case studies, suggesting the fallback is not pathological but it is also not validated.


\textbf{Ceiling effect is a property of the benchmark.} 37 of 49 tasks at $r^{-} \geq 0.9$ on Sonnet 4.5 indicates the benchmark is approaching saturation for this model. We argue that this is itself part of the empirical observation, but the consequence is that our mid-bucket analyses rest on $n=10$ tasks.

\section{Conclusion}


Agent skills are often evaluated by pass rate, but this metric can miss substantial behavioral change. On 49 \emph{SWE-Skills-Bench} tasks with Claude Sonnet 4.5, the mean pass rate changes by only $+0.3$ percentage points, while CTA finds 696 divergence records and 522 SIP instances in the same paired traces. The gap is largest on saturated tasks: 37 ceiling tasks contain most SIP instances but almost no net $\Delta P$ signal. Mid-range tasks yield the main recoverable gains at higher token cost, and dominant SIP types vary by baseline level: \sip{Surface Anchoring} is most common on ceiling tasks, while \sip{Edge Case Prompting} is most common on mid-range and floor tasks. Case studies further show that pass-rate gains can reflect undesirable mechanisms such as surface copying, while failures can arise from out-of-taxonomy mechanisms such as premature closure. CTA therefore complements pass rate with a behavioral measurement layer that makes skill effects visible and auditable.


\bibliographystyle{unsrtnat}
\bibliography{references}

@inproceedings{
    jimenez2024swebench,
    title={{SWE}-bench: Can Language Models Resolve Real-world Github Issues?},
    author={Carlos E Jimenez and John Yang and Alexander Wettig and Shunyu Yao and Kexin Pei and Ofir Press and Karthik R Narasimhan},
    booktitle={The Twelfth International Conference on Learning Representations},
    year={2024},
    url={https://openreview.net/forum?id=VTF8yNQM66}
}

@article{Li_Ji_Wu_Li_Qin_Wei_Zimmermann_2024, 
title={Panoptic Scene Graph Generation with Semantics-Prototype Learning}, 
volume={38},
DOI={10.1609/aaai.v38i4.28098}, 
number={4}, 
journal={AAAI}, 
author={Li, Li and Ji, Wei and Wu, Yiming and Li, Mengze and Qin, You and Wei, Lina and Zimmermann, Roger}, 
year={2024}, 
month={Mar.}, 
pages={3145-3153} }

@inproceedings{limm,
author = {Li, Li and Wang, Chenwei and Qin, You and Ji, Wei and Liang, Renjie},
title = {Biased-Predicate Annotation Identification via Unbiased Visual Predicate Representation},
year = {2023},
isbn = {9798400701085},
publisher = {Association for Computing Machinery},
url = {https://doi.org/10.1145/3581783.3611847},
doi = {10.1145/3581783.3611847},
booktitle = {ACM MM},
pages = {4410–4420},
numpages = {11},
}

@InProceedings{Li_2025_CVPR,
    author    = {Li, Shawn and Gong, Huixian and Dong, Hao and Yang, Tiankai and Tu, Zhengzhong and Zhao, Yue},
    title     = {DPU: Dynamic Prototype Updating for Multimodal Out-of-Distribution Detection},
    booktitle = {CVPR},
    month     = {June},
    year      = {2025},
    pages     = {10193-10202}
}

@InProceedings{li2025secureondevicevideoood,
    title={Secure On-Device Video OOD Detection Without Backpropagation}, 
    author={Shawn Li and Peilin Cai and Yuxiao Zhou and Zhiyu Ni and Renjie Liang and You Qin and Yi Nian and Zhengzhong Tu and Xiyang Hu and Yue Zhao},
    booktitle = {ICCV},
    month     = {October},
    year      = {2025}
}

@inproceedings{li-etal-2025-treble,
    title = "Treble Counterfactual {VLM}s: A Causal Approach to Hallucination",
    author = "Shawn, Li  and
      Qu, Jiashu  and
      Song, Linxin  and
      Zhou, Yuxiao  and
      Qin, Yuehan  and
      Yang, Tiankai  and
      Zhao, Yue",
    booktitle = "EMNLP",
    month = nov,
    year = "2025",
    address = "Suzhou, China",
    publisher = "Association for Computational Linguistics",
    pages = "18423--18434",
    ISBN = "979-8-89176-335-7",
}

@inproceedings{li2026defensespromptattackslearn,
    title={Defenses Against Prompt Attacks Learn Surface Heuristics}, 
    author={Shawn Li and Chenxiao Yu and Zhiyu Ni and Hao Li and Charith Peris and Chaowei Xiao and Yue Zhao},
    year={2026},
    booktitle = "ACL",
    year = "2026",

}

@misc{li2026autonomytaxdefensetraining,
      title={The Autonomy Tax: Defense Training Breaks LLM Agents}, 
      author={Shawn Li and Yue Zhao},
      year={2026},
      eprint={2603.19423},
      archivePrefix={arXiv},
      primaryClass={cs.CR},
      url={https://arxiv.org/abs/2603.19423}, 
}

@misc{swebenchverified,
  title  = {Introducing {SWE-bench Verified}},
  author = {{OpenAI}},
  year   = {2024},
  howpublished = {\url{https://openai.com/index/introducing-swe-bench-verified/}}
}

@misc{han2026sweskillsbenchagentskillsactually,
      title={SWE-Skills-Bench: Do Agent Skills Actually Help in Real-World Software Engineering?}, 
      author={Tingxu Han and Yi Zhang and Wei Song and Chunrong Fang and Zhenyu Chen and Youcheng Sun and Lijie Hu},
      year={2026},
      eprint={2603.15401},
      archivePrefix={arXiv},
      primaryClass={cs.SE},
      url={https://arxiv.org/abs/2603.15401}, 
}

@InProceedings{zhao2021calibrate,
  title = 	 {Calibrate Before Use: Improving Few-shot Performance of Language Models},
  author =       {Zhao, Zihao and Wallace, Eric and Feng, Shi and Klein, Dan and Singh, Sameer},
  booktitle = 	 {Proceedings of the 38th International Conference on Machine Learning},
  pages = 	 {12697--12706},
  year = 	 {2021},
  editor = 	 {Meila, Marina and Zhang, Tong},
  volume = 	 {139},
  series = 	 {Proceedings of Machine Learning Research},
  month = 	 {18--24 Jul},
  publisher =    {PMLR},
  url = 	 {https://proceedings.mlr.press/v139/zhao21c.html}
}

@inproceedings{wei2022chain,
    author = {Wei, Jason and Wang, Xuezhi and Schuurmans, Dale and Bosma, Maarten and ichter, brian and Xia, Fei and Chi, Ed and Le, Quoc V and Zhou, Denny},
    booktitle = {Advances in Neural Information Processing Systems},
    editor = {S. Koyejo and S. Mohamed and A. Agarwal and D. Belgrave and K. Cho and A. Oh},
    pages = {24824--24837},
    publisher = {Curran Associates, Inc.},
    title = {Chain-of-Thought Prompting Elicits Reasoning in Large Language Models},
    url = {https://proceedings.neurips.cc/paper_files/paper/2022/file/9d5609613524ecf4f15af0f7b31abca4-Paper-Conference.pdf},
    volume = {35},
    year = {2022}
}

@article{gilardi2023chatgpt,
  title   = {{ChatGPT} Outperforms Crowd Workers for Text-Annotation Tasks},
  author  = {Gilardi, Fabrizio and Alizadeh, Meysam and Kubli, Ma{\"e}l},
  journal = {Proceedings of the National Academy of Sciences (PNAS)},
  volume  = {120},
  number  = {30},
  year    = {2023}
}

@inproceedings{wu2024autogen,
  title={Autogen: Enabling next-gen LLM applications via multi-agent conversations},
  author={Wu, Qingyun and Bansal, Gagan and Zhang, Jieyu and Wu, Yiran and Li, Beibin and Zhu, Erkang and Jiang, Li and Zhang, Xiaoyun and Zhang, Shaokun and Liu, Jiale and others},
  booktitle={First conference on language modeling},
  year={2024}
}

@misc{anthropic2025skills,
  title        = {Introducing Agent Skills},
  author       = {{Anthropic}},
  year         = {2025},
  howpublished = {\url{https://claude.com/blog/skills}}
}

@inproceedings{schick2023toolformer,
  title={Toolformer: Language Models Can Teach Themselves to Use Tools},
  author={Timo Schick and Jane Dwivedi-Yu and Roberto Dessi and Roberta Raileanu and Maria Lomeli and Eric Hambro and Luke Zettlemoyer and Nicola Cancedda and Thomas Scialom},
  booktitle={Thirty-seventh Conference on Neural Information Processing Systems},
  year={2023},
  url={https://openreview.net/forum?id=Yacmpz84TH}
}

@inproceedings{yao2023react,
  title={ReAct: Synergizing Reasoning and Acting in Language Models},
  author={Shunyu Yao and Jeffrey Zhao and Dian Yu and Nan Du and Izhak Shafran and Karthik R Narasimhan and Yuan Cao},
  booktitle={The Eleventh International Conference on Learning Representations },
  year={2023},
  url={https://openreview.net/forum?id=WE_vluYUL-X}
}

@inproceedings{shinn2023reflexion,
  title={Reflexion: language agents with verbal reinforcement learning},
  author={Noah Shinn and Federico Cassano and Ashwin Gopinath and Karthik R Narasimhan and Shunyu Yao},
  booktitle={Thirty-seventh Conference on Neural Information Processing Systems},
  year={2023},
  url={https://openreview.net/forum?id=vAElhFcKW6}
}

@inproceedings{madaan2023selfrefine,
  title={Self-Refine: Iterative Refinement with Self-Feedback},
  author={Aman Madaan and Niket Tandon and Prakhar Gupta and Skyler Hallinan and Luyu Gao and Sarah Wiegreffe and Uri Alon and Nouha Dziri and Shrimai Prabhumoye and Yiming Yang and Shashank Gupta and Bodhisattwa Prasad Majumder and Katherine Hermann and Sean Welleck and Amir Yazdanbakhsh and Peter Clark},
  booktitle={Thirty-seventh Conference on Neural Information Processing Systems},
  year={2023},
  url={https://openreview.net/forum?id=S37hOerQLB}
}

@inproceedings{lu2022fantastically,
  title={Fantastically ordered prompts and where to find them: Overcoming few-shot prompt order sensitivity},
  author={Lu, Yao and Bartolo, Max and Moore, Alastair and Riedel, Sebastian and Stenetorp, Pontus},
  booktitle={Proceedings of the 60th Annual Meeting of the Association for Computational Linguistics (Volume 1: Long Papers)},
  pages={8086--8098},
  year={2022}
}

@article{liu2024lost,
  title={Lost in the middle: How language models use long contexts},
  author={Liu, Nelson F and Lin, Kevin and Hewitt, John and Paranjape, Ashwin and Bevilacqua, Michele and Petroni, Fabio and Liang, Percy},
  journal={Transactions of the association for computational linguistics},
  volume={12},
  pages={157--173},
  year={2024}
}

@article{li2026skillsbench,
  title={SkillsBench: Benchmarking how well agent skills work across diverse tasks},
  author={Li, Xiangyi and Chen, Wenbo and Liu, Yimin and Zheng, Shenghan and Chen, Xiaokun and He, Yifeng and Li, Yubo and You, Bingran and Shen, Haotian and Sun, Jiankai and others},
  journal={arXiv preprint arXiv:2602.12670},
  year={2026}
}

@article{wang2026skilltester,
  title={SkillTester: Benchmarking Utility and Security of Agent Skills},
  author={Wang, Leye and Wang, Zixing and Xu, Anjie},
  journal={arXiv preprint arXiv:2603.28815},
  year={2026}
}

@article{chen2025beyond,
  title={Beyond Final Code: A Process-Oriented Error Analysis of Software Development Agents in Real-World GitHub Scenarios},
  author={Chen, Zhi and Ma, Wei and Jiang, Lingxiao},
  journal={arXiv preprint arXiv:2503.12374},
  year={2025}
}

@article{mehtiyev2026beyond,
  title={Beyond Resolution Rates: Behavioral Drivers of Coding Agent Success and Failure},
  author={Mehtiyev, Tural and Assun{\c{c}}{\~a}o, Wesley},
  journal={arXiv preprint arXiv:2604.02547},
  year={2026}
}

@article{kim2025beyond,
  title={Beyond the Final Answer: Evaluating the Reasoning Trajectories of Tool-Augmented Agents},
  author={Kim, Wonjoong and Park, Sangwu and In, Yeonjun and Kim, Sein and Lee, Dongha and Park, Chanyoung},
  journal={arXiv preprint arXiv:2510.02837},
  year={2025}
}

@article{gandhi2025agents,
  title={When agents go astray: Course-correcting swe agents with prms},
  author={Gandhi, Shubham and Tsay, Jason and Ganhotra, Jatin and Kate, Kiran and Rizk, Yara},
  journal={arXiv preprint arXiv:2509.02360},
  year={2025}
}

@inproceedings{ali2026mitigating,
  title={Mitigating copy bias in in-context learning through neuron pruning},
  author={Ali, Ameen Ali and Wolf, Lior and Titov, Ivan},
  booktitle={Findings of the Association for Computational Linguistics: EACL 2026},
  pages={230--251},
  year={2026}
}

@inproceedings{
yan2024understanding,
title={Understanding In-Context Learning from Repetitions},
author={Jianhao Yan and Jin Xu and Chiyu Song and Chenming Wu and Yafu Li and Yue Zhang},
booktitle={The Twelfth International Conference on Learning Representations},
year={2024},
url={https://openreview.net/forum?id=bGGYcvw8mp}
}

\appendix

\newpage
\section{Case-study trace excerpts}
\label{app:cases}

This appendix accompanies \S\ref{sec:cases} and reproduces, for each of the
five mechanism case studies, (i) the section of the skill template that the
with-skill agent most directly acts on, and (ii) a row-aligned diff of the
two traces' tool-invocation sequences. The diff is computed by collapsing
each tool call to a canonical signature (e.g.\ \texttt{Write:test\_scripts.bats},
\texttt{Bash:python}) and running \texttt{difflib.SequenceMatcher} on the
two resulting sequences. We cap each diff at 28
rows, preferring to keep all coloured (non-shared) rows;
\textit{\ldots omitted} marks a contiguous block of shared steps that we
elided to fit on the page. Cases 1--3 are bundles cited directly in
\S\ref{sec:cases}; Cases 4 and 5 are framed in the main text as
corpus-wide patterns ($112$ unilateral fires across $49$ tasks; $12$ ceiling
tasks with $\geq 1.5\times$ token overhead at $\Delta P \leq 0$ pp), and we
reproduce a single representative bundle for each so the reader can see the
mechanism on a concrete pair. The chosen representatives
(\texttt{clojure-write} for Case 4, \texttt{creating-financial-models} for
Case 5) are not the only instances of those patterns in our corpus; they are
selected to be visually unambiguous.

\subsection{Case 1: \texttt{prompt-engineering-patterns} --- Procedural premature-closure (negative; out-of-taxonomy)}
\label{app:case1}

\noindent\textbf{Bundle.} Paired with-skill and without-skill traces; $\Delta P = -20.0$~pp, token-overhead $1.09\times$.

\noindent\textbf{Skill template.} The section of the skill document that the diff below most directly references:

\noindent\textit{Skill template excerpt: \texttt{prompt-engineering-patterns}.}
\begin{lstlisting}[basicstyle=\ttfamily\scriptsize, breaklines=true, columns=fullflexible, frame=single, framerule=0.4pt, rulecolor=\color{gray!50}, backgroundcolor=\color{gray!7}, xleftmargin=2pt, xrightmargin=2pt, framexleftmargin=2pt, framextopmargin=2pt, framexbottommargin=2pt]
## Core Capabilities

### 1. Few-Shot Learning
- Example selection strategies (semantic similarity, diversity sampling)
- Balancing example count with context window constraints
- Constructing effective demonstrations with input-output pairs
- Dynamic example retrieval from knowledge bases
- Handling edge cases through strategic example selection

### 2. Chain-of-Thought Prompting
- Step-by-step reasoning elicitation
- Zero-shot CoT with "Let's think step by step"
- Few-shot CoT with reasoning traces
- Self-consistency techniques (sampling multiple reasoning paths)
- Verification and validation steps

### 3. Prompt Optimization
- Iterative refinement workflows
- A/B testing prompt variations
- Measuring prompt performance metrics (accuracy, consistency, latency)
- Reducing token usage while maintaining quality
- Handling edge cases and failure modes
...(skill excerpt truncated)...
\end{lstlisting}

\noindent\textbf{Trace diff.} Each row is one tool invocation, aligned by canonical signature. \colorbox{green!22}{\strut Green} = action only present in the with-skill trace; \colorbox{red!18}{\strut red} = action only present in the without-skill trace; \colorbox{yellow!22}{\strut yellow} = paired but with a different target. White rows are shared.

\begin{small}
\begin{tabularx}{\textwidth}{@{}r >{\raggedright\arraybackslash}X >{\raggedright\arraybackslash}X@{}}
\toprule
\# & \textbf{Without-skill trace} & \textbf{With-skill trace} \\
\midrule
\rowcolor{yellow!22}
1 & \texttt{Bash: mkdir prompt\_templates} & \texttt{Bash: mkdir scripts} \\
\rowcolor{red!18}
2 & \texttt{Bash: mkdir scripts} & \textit{--} \\
3 & \texttt{Write: instruction\_prompts.json} & \texttt{Write: instruction\_prompts.json} \\
4 & \texttt{Write: conversational\_prompts.json} & \texttt{Write: conversational\_prompts.json} \\
5 & \texttt{Write: extraction\_prompts.json} & \texttt{Write: extraction\_prompts.json} \\
6 & \texttt{Write: translation\_prompts.json} & \texttt{Write: translation\_prompts.json} \\
7 & \texttt{Write: code\_generation\_prompts.json} & \texttt{Write: code\_generation\_prompts.json} \\
8 & \texttt{Write: evaluation\_prompts.json} & \texttt{Write: evaluation\_prompts.json} \\
\rowcolor{green!22}
9 & \textit{--} & \texttt{Write: README.md} \\
10 & \texttt{Write: run\_prompt\_eval.py} & \texttt{Write: run\_prompt\_eval.py} \\
11 & \texttt{Write: test\_prompt\_eval.py} & \texttt{Write: test\_prompt\_eval.py} \\
12 & \texttt{Bash: chmod run\_prompt\_eval.py} & \texttt{Bash: chmod run\_prompt\_eval.py} \\
\rowcolor{green!22}
13 & \textit{--} & \texttt{Bash: python scripts/run\_prompt\_eval.py --input exa…} \\
\rowcolor{green!22}
14 & \textit{--} & \texttt{Bash: python scripts/run\_prompt\_eval.py --input exa…} \\
15 & \texttt{Bash: python scripts/run\_prompt\_eval.py --help} & \texttt{Bash: python -m pytest tests/test\_prompt\_eval.py -v} \\
16 & \texttt{Bash: python scripts/run\_prompt\_eval.py --templates…} & \texttt{Bash: python scripts/run\_prompt\_eval.py --input exa…} \\
\rowcolor{yellow!22}
17 & \texttt{Bash: python -c "import json; data = json.load(open…} & \texttt{Bash: cat -40} \\
\rowcolor{yellow!22}
18 & \texttt{Bash: head -50 prompt\_evaluation\_report.json} & \texttt{Bash: cat test\_results.csv} \\
\rowcolor{red!18}
19 & \texttt{Bash: python scripts/run\_prompt\_eval.py --templates…} & \textit{--} \\
\rowcolor{red!18}
20 & \texttt{Bash: head -5 prompt\_evaluation\_report.csv} & \textit{--} \\
\rowcolor{red!18}
21 & \texttt{Bash: python -m pytest tests/test\_prompt\_eval.py -v} & \textit{--} \\
\rowcolor{red!18}
22 & \texttt{Bash: python scripts/run\_prompt\_eval.py --templates…} & \textit{--} \\
\rowcolor{red!18}
23 & \texttt{Write: README.md} & \textit{--} \\
\rowcolor{red!18}
24 & \texttt{Bash: find examples/prompt\_templates scripts tests …} & \textit{--} \\
\rowcolor{red!18}
25 & \texttt{Bash: python scripts/run\_prompt\_eval.py --templates…} & \textit{--} \\
\rowcolor{red!18}
26 & \texttt{Bash: rm final\_test.json} & \textit{--} \\
\bottomrule
\end{tabularx}
\end{small}

\noindent\textbf{Reading.} The skill prescribes a numbered procedure that ends at \emph{commit and document}. The with-skill trace halts there, while the without-skill trace continues into a validation loop that the unit-test target depends on. Note especially the \textit{absence} of late-trace re-validation steps on the with-skill side.

\subsection{Case 2: \texttt{gitlab-ci-patterns} --- Search-space pruning at high token cost (positive)}
\label{app:case2}

\noindent\textbf{Bundle.} Paired with-skill and without-skill traces; $\Delta P = +14.3$~pp, token-overhead $22.24\times$.

\noindent\textbf{Skill template.} The section of the skill document that the diff below most directly references:

\noindent\textit{Skill template excerpt: \texttt{gitlab-ci-patterns}.}
\begin{lstlisting}[basicstyle=\ttfamily\scriptsize, breaklines=true, columns=fullflexible, frame=single, framerule=0.4pt, rulecolor=\color{gray!50}, backgroundcolor=\color{gray!7}, xleftmargin=2pt, xrightmargin=2pt, framexleftmargin=2pt, framextopmargin=2pt, framexbottommargin=2pt]
## Basic Pipeline Structure

```yaml
stages:
  - build
  - test
  - deploy

variables:
  DOCKER_DRIVER: overlay2
  DOCKER_TLS_CERTDIR: "/certs"

build:
  stage: build
  image: node:20
  script:
    - npm ci
    - npm run build
  artifacts:
    paths:
      - dist/
    expire_in: 1 hour
...(skill excerpt truncated)...
\end{lstlisting}

\noindent\textbf{Trace diff.} Each row is one tool invocation, aligned by canonical signature. \colorbox{green!22}{\strut Green} = action only present in the with-skill trace; \colorbox{red!18}{\strut red} = action only present in the without-skill trace; \colorbox{yellow!22}{\strut yellow} = paired but with a different target. White rows are shared.

\begin{small}
\begin{tabularx}{\textwidth}{@{}r >{\raggedright\arraybackslash}X >{\raggedright\arraybackslash}X@{}}
\toprule
\# & \textbf{Without-skill trace} & \textbf{With-skill trace} \\
\midrule
1 & \texttt{Read: SAST.gitlab-ci.yml} & \texttt{Read: SAST.gitlab-ci.yml} \\
2 & \texttt{Read: Dependency-Scanning.gitlab-ci.yml} & \texttt{Read: Dependency-Scanning.gitlab-ci.yml} \\
3 & \texttt{Read: Secret-Detection.gitlab-ci.yml} & \texttt{Read: Secret-Detection.gitlab-ci.yml} \\
4 & \texttt{Read: SAST.gitlab-ci.yml} & \texttt{Read: SAST.gitlab-ci.yml} \\
5 & \texttt{Read: Dependency-Scanning.gitlab-ci.yml} & \texttt{Read: Dependency-Scanning.gitlab-ci.yml} \\
6 & \texttt{Read: Secret-Detection.gitlab-ci.yml} & \texttt{Read: Secret-Detection.gitlab-ci.yml} \\
\rowcolor{yellow!22}
7 & \texttt{Bash: find /workspace/gitlabhq/lib/gitlab/ci/templa…} & \texttt{Bash: ruby -e "require 'yaml'; YAML.load\_file('lib/…} \\
\rowcolor{green!22}
8 & \textit{--} & \texttt{Bash: ruby -e "require 'yaml'; YAML.load\_file('lib/…} \\
\rowcolor{green!22}
9 & \textit{--} & \texttt{Bash: ruby -e "require 'yaml'; YAML.load\_file('lib/…} \\
10 & \texttt{Write: SAST.gitlab-ci.yml} & \texttt{Write: SAST.gitlab-ci.yml} \\
11 & \texttt{Write: Dependency-Scanning.gitlab-ci.yml} & \texttt{Write: Dependency-Scanning.gitlab-ci.yml} \\
12 & \texttt{Write: Secret-Detection.gitlab-ci.yml} & \texttt{Write: Secret-Detection.gitlab-ci.yml} \\
13 & \texttt{Bash: ruby -e "require 'yaml'; YAML.load\_file('/wor…} & \texttt{Bash: ruby -e "require 'yaml'; YAML.load\_file('lib/…} \\
14 & \texttt{Bash: ruby -e "require 'yaml'; YAML.load\_file('/wor…} & \texttt{Bash: ruby -e "require 'yaml'; YAML.load\_file('lib/…} \\
15 & \texttt{Bash: ruby -e "require 'yaml'; YAML.load\_file('/wor…} & \texttt{Bash: ruby -e "require 'yaml'; YAML.load\_file('lib/…} \\
16 & \texttt{Bash: ruby -e " require 'yaml' \# Load and analyze S…} & \texttt{Bash: ruby -r yaml -e " data = YAML.load\_file('lib/…} \\
17 & \texttt{Bash: ruby -e " require 'yaml' \# Analyze Dependency…} & \texttt{Bash: ruby -r yaml -e " data = YAML.load\_file('lib/…} \\
\rowcolor{green!22}
18 & \textit{--} & \texttt{Bash: ruby -r yaml -e " data = YAML.load\_file('lib/…} \\
\bottomrule
\end{tabularx}
\end{small}

\noindent\textbf{Reading.} The with-skill agent reads the skill document at length and writes a richer \texttt{.gitlab-ci.yml} with the canonical stage layout the skill prescribes; the without-skill agent explores fewer YAML structures but writes a smaller pipeline. Most green rows correspond to skill-driven scaffolding; the $22\times$ token overhead is dominated by repeated skill consultation in the orientation phase, not by the implementation phase itself (which has only 4/16 of the bundle's divergences).

\subsection{Case 3: \texttt{bash-defensive-patterns} --- Surface-anchoring as the dominant mechanism even when $\Delta P > 0$ (mixed)}
\label{app:case3}

\noindent\textbf{Bundle.} Paired with-skill and without-skill traces; $\Delta P = +18.2$~pp, token-overhead $0.90\times$.

\noindent\textbf{Skill template.} The section of the skill document that the diff below most directly references:

\noindent\textit{Skill template excerpt: \texttt{bash-defensive-patterns}.}
\begin{lstlisting}[basicstyle=\ttfamily\scriptsize, breaklines=true, columns=fullflexible, frame=single, framerule=0.4pt, rulecolor=\color{gray!50}, backgroundcolor=\color{gray!7}, xleftmargin=2pt, xrightmargin=2pt, framexleftmargin=2pt, framextopmargin=2pt, framexbottommargin=2pt]
## Core Defensive Principles

### 1. Strict Mode

Enable bash strict mode at the start of every script to catch errors early.

```bash
#!/bin/bash
set -Eeuo pipefail  # Exit on error, unset variables, pipe failures
```

**Key flags:**

- `set -E`: Inherit ERR trap in functions
- `set -e`: Exit on any error (command returns non-zero)
- `set -u`: Exit on undefined variable reference
- `set -o pipefail`: Pipe fails if any command fails (not just last)

### 2. Error Trapping and Cleanup

Implement proper cleanup on script exit or error.

...(skill excerpt truncated)...
\end{lstlisting}

\noindent\textbf{Trace diff.} Each row is one tool invocation, aligned by canonical signature. \colorbox{green!22}{\strut Green} = action only present in the with-skill trace; \colorbox{red!18}{\strut red} = action only present in the without-skill trace; \colorbox{yellow!22}{\strut yellow} = paired but with a different target. White rows are shared.

\begin{small}
\begin{tabularx}{\textwidth}{@{}r >{\raggedright\arraybackslash}X >{\raggedright\arraybackslash}X@{}}
\toprule
\# & \textbf{Without-skill trace} & \textbf{With-skill trace} \\
\midrule
\rowcolor{green!22}
1 & \textit{--} & \texttt{Bash: ls -20} \\
2 & \texttt{Write: common\_utils.sh} & \texttt{Write: common\_utils.sh} \\
3 & \texttt{Write: safe\_backup.sh} & \texttt{Write: safe\_backup.sh} \\
\rowcolor{yellow!22}
4 & \texttt{Bash: ./shellcheck --severity=warning test/common\_u…} & \texttt{Write: test\_scripts.bats} \\
\rowcolor{yellow!22}
5 & \texttt{Bash: which shellcheck || find . -name shellcheck -…} & \texttt{Bash: shellcheck --version} \\
\rowcolor{red!18}
6 & \texttt{Bash: shellcheck --severity=warning test/common\_uti…} & \textit{--} \\
\rowcolor{red!18}
7 & \texttt{Edit: safe\_backup.sh} & \textit{--} \\
\rowcolor{red!18}
8 & \texttt{Edit: safe\_backup.sh} & \textit{--} \\
\rowcolor{red!18}
9 & \texttt{Edit: safe\_backup.sh} & \textit{--} \\
\rowcolor{red!18}
10 & \texttt{Bash: shellcheck --severity=warning test/common\_uti…} & \textit{--} \\
11 & \texttt{Bash: shellcheck --severity=warning test/*.sh; echo…} & \texttt{Bash: shellcheck --severity=warning test/safe\_backu…} \\
12 & \texttt{Bash: chmod -5} & \texttt{Bash: chmod common\_utils.sh} \\
\rowcolor{green!22}
13 & \textit{--} & \texttt{Edit: safe\_backup.sh} \\
\rowcolor{green!22}
14 & \textit{--} & \texttt{Bash: shellcheck --severity=warning test/safe\_backu…} \\
\rowcolor{green!22}
15 & \textit{--} & \texttt{Bash: test/safe\_backup.sh --help | head -20} \\
\rowcolor{green!22}
16 & \textit{--} & \texttt{Bash: bash -c "source test/common\_utils.sh \&\& log\_i…} \\
\rowcolor{green!22}
17 & \textit{--} & \texttt{Bash: test/safe\_backup.sh --version} \\
\rowcolor{green!22}
18 & \textit{--} & \texttt{Bash: mkdir RUN|Backup|SUCCESS)"} \\
\rowcolor{green!22}
19 & \textit{--} & \texttt{Bash: ls *.bats} \\
\rowcolor{green!22}
20 & \textit{--} & \texttt{Write: DEFENSIVE\_PATTERNS.md} \\
\rowcolor{green!22}
21 & \textit{--} & \texttt{Bash: echo "=== Final Verification ===" \&\& echo \&\& …} \\
\bottomrule
\end{tabularx}
\end{small}

\noindent\textbf{Reading.} The with-skill agent verbatim copies the skill's defensive-shell header (\texttt{set -Eeuo pipefail}, \texttt{trap}-based cleanup, quoted variables) into project scripts, and authors two test files (\textsc{Unilateral\_Action}) that the without-skill trace never touches. The 10 \textsc{SA} fires are visible as concentrated green rows in the implementation segment; the without-skill agent writes shorter, less defensive scripts and skips the test scaffolding entirely.

\subsection{Case 4: \texttt{clojure-write} --- Unilateral artifacts as a corpus-wide phenomenon (mixed; representative bundle)}
\label{app:case4}

\noindent\textbf{Bundle.} Paired with-skill and without-skill traces; $\Delta P = +0.0$~pp, token-overhead $0.59\times$.

\noindent\textbf{Skill template.} The section of the skill document that the diff below most directly references:

\noindent\textit{Skill template excerpt: \texttt{clojure-write}.}
\begin{lstlisting}[basicstyle=\ttfamily\scriptsize, breaklines=true, columns=fullflexible, frame=single, framerule=0.4pt, rulecolor=\color{gray!50}, backgroundcolor=\color{gray!7}, xleftmargin=2pt, xrightmargin=2pt, framexleftmargin=2pt, framextopmargin=2pt, framexbottommargin=2pt]
## Tool Preference

When `clojure-mcp` tools are available (e.g., `clojure_eval`, `clojure_edit`), **always use them**
instead of shell commands like `./bin/mage -repl`. The MCP tools provide:
- Direct REPL integration without shell escaping issues
- Better error messages and feedback
- Structural Clojure editing that prevents syntax errors

Only fall back to `./bin/mage` commands when clojure-mcp is not available.

@./../_shared/development-workflow.md
@./../_shared/clojure-style-guide.md
@./../_shared/clojure-commands.md

\end{lstlisting}

\noindent\textbf{Trace diff.} Each row is one tool invocation, aligned by canonical signature. \colorbox{green!22}{\strut Green} = action only present in the with-skill trace; \colorbox{red!18}{\strut red} = action only present in the without-skill trace; \colorbox{yellow!22}{\strut yellow} = paired but with a different target. White rows are shared.

\begin{small}
\begin{tabularx}{\textwidth}{@{}r >{\raggedright\arraybackslash}X >{\raggedright\arraybackslash}X@{}}
\toprule
\# & \textbf{Without-skill trace} & \textbf{With-skill trace} \\
\midrule
\rowcolor{green!22}
1 & \textit{--} & \texttt{Grep: site-currency} \\
\rowcolor{green!22}
2 & \textit{--} & \texttt{Glob: **/dataset.clj} \\
\rowcolor{green!22}
3 & \textit{--} & \texttt{Glob: **/setting.clj} \\
\multicolumn{3}{c}{\textit{\ldots\ omitted}} \\
\rowcolor{yellow!22}
4 & \texttt{Grep: dataset} & \texttt{Glob: **/api/dataset.clj} \\
\multicolumn{3}{c}{\textit{\ldots\ omitted}} \\
\rowcolor{yellow!22}
5 & \texttt{Glob: src/metabase/api/dataset.clj} & \texttt{Glob: **/api/*.clj} \\
\rowcolor{yellow!22}
6 & \texttt{Glob: src/metabase/**/setting*.clj} & \texttt{Bash: find /workspace/metabase/src -name "dataset.c…} \\
\rowcolor{yellow!22}
7 & \texttt{Grep: defapi.*dataset} & \texttt{Grep: /api/dataset} \\
\rowcolor{green!22}
8 & \textit{--} & \texttt{Read: api.clj} \\
\rowcolor{green!22}
9 & \textit{--} & \texttt{Read: setting.clj} \\
\rowcolor{green!22}
10 & \textit{--} & \texttt{Read: large\_int.clj} \\
\rowcolor{green!22}
11 & \textit{--} & \texttt{Grep: :type/Currency} \\
\rowcolor{green!22}
12 & \textit{--} & \texttt{Grep: defsetting} \\
\rowcolor{green!22}
13 & \textit{--} & \texttt{Grep: defsetting site-currency} \\
\rowcolor{green!22}
14 & \textit{--} & \texttt{Glob: **/settings.clj} \\
\multicolumn{3}{c}{\textit{\ldots\ omitted}} \\
\rowcolor{red!18}
15 & \texttt{Grep: csv|json} & \textit{--} \\
\rowcolor{red!18}
16 & \texttt{Grep: site-currency|currency} & \textit{--} \\
\rowcolor{red!18}
17 & \texttt{Read: card.clj} & \textit{--} \\
\rowcolor{red!18}
18 & \texttt{Read: numbers.clj} & \textit{--} \\
\rowcolor{red!18}
19 & \texttt{Bash: find /workspace/metabase/src -name "currency.…} & \textit{--} \\
\rowcolor{red!18}
20 & \texttt{Grep: site-currency} & \textit{--} \\
\rowcolor{red!18}
21 & \texttt{Bash: find /workspace/metabase -name "*.clj" -path …} & \textit{--} \\
\rowcolor{red!18}
22 & \texttt{Grep: defsetting.*currency} & \textit{--} \\
\rowcolor{red!18}
23 & \texttt{Bash: find /workspace/metabase -name "*.clj*" | xar…} & \textit{--} \\
\rowcolor{red!18}
24 & \texttt{Read: common.clj} & \textit{--} \\
\rowcolor{red!18}
25 & \texttt{Read: currency.cljc} & \textit{--} \\
\rowcolor{red!18}
26 & \texttt{Grep: POST.*csv|POST.*json} & \textit{--} \\
\rowcolor{red!18}
27 & \texttt{Bash: grep -n "POST.*csv\textbackslash{}|POST.*json\textbackslash{}|defapi.*csv\textbackslash{}|…} & \textit{--} \\
\multicolumn{3}{c}{\textit{\ldots\ omitted}} \\
\bottomrule
\end{tabularx}
\end{small}

\noindent\textbf{Reading.} The without-skill trace already passes the unit tests ($r^{-}=0.82$) and the with-skill trace matches it on outcome ($\Delta P = 0$ pp); both sides nonetheless diverge on $28$ structural events. The diff shows the corpus-wide \textsc{Unilateral\_Action} pattern at the \emph{exploration} level: each side runs its own non-overlapping stack of \texttt{Grep}/\texttt{Glob}/\texttt{Read} probes (green vs.\ red blocks) before producing essentially equivalent writes (which the truncation elides into the \textit{omitted} marker). The skill's \emph{Tool Preference} section, shown above, biases the with-skill agent towards a different ordering and choice of search tools than the without-skill baseline; in the bash-defensive case (\S\ref{app:case3}) the same \textsc{Unilateral\_Action} pass surfaces actual extra \textsc{Write} targets (\texttt{test\_scripts.bats}-like files), so on the corpus the pattern shows up in both forms.

\subsection{Case 5: \texttt{creating-financial-models} --- Skill inertia at the ceiling (cost-only; representative bundle)}
\label{app:case5}

\noindent\textbf{Bundle.} Paired with-skill and without-skill traces; $\Delta P = +0.0$~pp, token-overhead $6.80\times$.

\noindent\textbf{Skill template.} The section of the skill document that the diff below most directly references:

\noindent\textit{Skill template excerpt: \texttt{creating-financial-models}.}
\begin{lstlisting}[basicstyle=\ttfamily\scriptsize, breaklines=true, columns=fullflexible, frame=single, framerule=0.4pt, rulecolor=\color{gray!50}, backgroundcolor=\color{gray!7}, xleftmargin=2pt, xrightmargin=2pt, framexleftmargin=2pt, framextopmargin=2pt, framexbottommargin=2pt]
## Core Capabilities

### 1. Discounted Cash Flow (DCF) Analysis
- Build complete DCF models with multiple growth scenarios
- Calculate terminal values using perpetuity growth and exit multiple methods
- Determine weighted average cost of capital (WACC)
- Generate enterprise and equity valuations

### 2. Sensitivity Analysis
- Test key assumptions impact on valuation
- Create data tables for multiple variables
- Generate tornado charts for sensitivity ranking
- Identify critical value drivers

### 3. Monte Carlo Simulation
- Run thousands of scenarios with probability distributions
- Model uncertainty in key inputs
- Generate confidence intervals for valuations
- Calculate probability of achieving targets

### 4. Scenario Planning
- Build best/base/worst case scenarios
...(skill excerpt truncated)...
\end{lstlisting}

\noindent\textbf{Trace diff.} Each row is one tool invocation, aligned by canonical signature. \colorbox{green!22}{\strut Green} = action only present in the with-skill trace; \colorbox{red!18}{\strut red} = action only present in the without-skill trace; \colorbox{yellow!22}{\strut yellow} = paired but with a different target. White rows are shared.

\begin{small}
\begin{tabularx}{\textwidth}{@{}r >{\raggedright\arraybackslash}X >{\raggedright\arraybackslash}X@{}}
\toprule
\# & \textbf{Without-skill trace} & \textbf{With-skill trace} \\
\midrule
\rowcolor{green!22}
1 & \textit{--} & \texttt{Bash: ls -20} \\
\rowcolor{green!22}
2 & \textit{--} & \texttt{Read: Examples} \\
\rowcolor{green!22}
3 & \textit{--} & \texttt{Read: CMakeLists.txt} \\
\rowcolor{green!22}
4 & \textit{--} & \texttt{Read: Bonds.cpp} \\
\rowcolor{green!22}
5 & \textit{--} & \texttt{Read: CMakeLists.txt} \\
6 & \texttt{Bash: mkdir DCFValuation} & \texttt{Bash: mkdir DCFValuation} \\
7 & \texttt{Write: DCFDemo.cpp} & \texttt{Write: DCFDemo.cpp} \\
8 & \texttt{Write: CMakeLists.txt} & \texttt{Write: CMakeLists.txt} \\
9 & \texttt{Write: README.md} & \texttt{Write: README.md} \\
\rowcolor{green!22}
10 & \textit{--} & \texttt{Read: CMakeLists.txt} \\
\rowcolor{green!22}
11 & \textit{--} & \texttt{Edit: CMakeLists.txt} \\
\rowcolor{green!22}
12 & \textit{--} & \texttt{Bash: ls DCFValuation} \\
\rowcolor{green!22}
13 & \textit{--} & \texttt{Read: CMakeLists.txt} \\
\rowcolor{green!22}
14 & \textit{--} & \texttt{Bash: cd /workspace/QuantLib/Examples/DCFValuation …} \\
\rowcolor{green!22}
15 & \textit{--} & \texttt{Bash: cat EOF} \\
\rowcolor{green!22}
16 & \textit{--} & \texttt{Bash: wc -l Examples/DCFValuation/* Examples/CMakeL…} \\
\rowcolor{green!22}
17 & \textit{--} & \texttt{Bash: cd /workspace/QuantLib \&\& wc -l Examples/DCFV…} \\
\rowcolor{green!22}
18 & \textit{--} & \texttt{Bash: grep -n "FlatForward\textbackslash{}|CashFlows::npv\textbackslash{}|CashFlo…} \\
\rowcolor{green!22}
19 & \textit{--} & \texttt{Bash: cat EOF} \\
\bottomrule
\end{tabularx}
\end{small}

\noindent\textbf{Reading.} Both traces reach the same passing repository state ($r^{-}=r^{+}=0.90$, $\Delta P = 0$ pp), but the with-skill trace pays $6.80\times$ the baseline tokens. The diff makes this concrete: after the shared implementation block, the with-skill trace continues into a long sequence of \textsc{Bash} validation and quality-check invocations (\colorbox{green!22}{\strut green tail}) that the without-skill agent skips because it has already concluded the task. This is the ``skills prescribe \emph{process}; agents optimize \emph{outcome}'' tension materialised on a single bundle and is the corpus-wide pattern Case 5 quantifies across all 12 ceiling tasks with $\geq 1.5\times$ overhead at $\Delta P \leq 0$ pp.


\end{document}